\documentclass[conference]{IEEEtran}
\ifCLASSINFOpdf
\else
\fi
\usepackage{cite}
\usepackage{amsmath,amssymb,amsfonts}
\usepackage{algorithmic}
\usepackage{graphicx}
\usepackage{textcomp}
\usepackage{array}
\usepackage{booktabs} 
\usepackage{algorithm}
\usepackage{enumerate}

\begin{document}
%
\title{Chinese User Service Intention Classification Based on Hybrid Neural Network}

\author{\IEEEauthorblockN{Shengbin Jia and Yang Xiang}
\IEEEauthorblockA{College of Electronics and Information Engineering\\Tongji University\\Shanghai, China}}


%


\maketitle

\begin{abstract}
In order to satisfy the consumers' increasing personalized service demand, the Intelligent service has arisen. User service intention recognition is an important challenge for intelligent service system to provide precise service. It is difficult for the intelligent system to understand the semantics of user demand which leads to poor recognition effect, because of the noise in user requirement descriptions. Therefore, a hybrid neural network classification model based on BiLSTM and CNN is proposed to recognize users’ service intentions. The model can fuse the temporal semantics and spatial semantics of the user descriptions. The experimental results show that our model achieves a better effect compared with other models, reaching 0.94 on the F1 score.
\newline

\emph{Keywords-User service intention recognition; Neural network; Text classification; Intelligent service system}

\end{abstract}


%
\IEEEpeerreviewmaketitle

\section{Introduction}
The technologies of the mobile Internet, artificial intelligence, and the Internet of things are in the ascendant, and the ``Intelligent service''~\cite{Liu2018The,Poniszewska2018Endowing} has come into being, such as intelligent chat robot, personal affairs assistant. Every service request made by the user contains the user's demand intention. The intelligent service system provides personalized service for different intentions. Therefore, how to understand the user's service intentions correctly is a primary challenge when the intelligent system provides accurate service. 

User service intentions may involve multiple fields during service processing. Therefore, it is key for automatically identifying user service intentions to correctly classify user intentions into corresponding fields. For example, when a user makes a request to ``Help me write an email for ...", the system should automatically recognize it as the ``email" category, and subsequent service actions will be carried out in the ``email" area.

In the intelligent service system, user's service demand is usually described by natural language. Users often cannot describe their intentions properly. These descriptions are short, changeable in ways, serious colloquial, and even carry typos and other noises. The main problems of difficult identification of user service demand are embodied as follows:
\begin{itemize}
	\item Users' input is not standardized, input methods are diversified. Users use natural language query, or even non-standard natural language. For example, "nearby special hotel", "Shanghai hotel how to go" and so on.
	\item  The users' query terms show multiple intentions, such as whether the user searches for "apples" to refer to fruit or electronic products, which is difficult to confirm.
	\item Intention exists timeliness change, that is, as time goes on, the intention of some query words will change. For example, the HUAWEI P10 was launched in March 24th. The demand intent on March 21 should be to know more about the mobile phone, and the intention to buy the mobile phone after March 24 may be stronger.
	\item Data cold start problem, when user behavior data is few, it is very difficult to get the user's search intention accurately.
\end{itemize}

Therefore, there is a trouble to accurately understand the user's intention because of the lack of deep understanding of the text semantics~\cite{Fubo2015Consumption}. In the early stage, the identification of consumption intentions used complex artificial features, mainly based on template matching methods and traditional classification methods~\cite{Zhang2014Mining},~\cite{haffner2003optimizing}. In recent years, with the development of the neural network, many natural languages processing tasks, including service intention recognition task, have been further developed.

This paper presents a user service intention classification model based on a hybrid neural network, which is a multi-layer network consisting of bidirectional long and short-term memory (BiLSTM) and convolutional neural network (CNN). It utilizes BiLSTM to capture both the forward and backward context semantics of each word. Meanwhile, CNN is used to extract salient local features. 

\begin{figure*}
	\centering
	\includegraphics[width=7in]{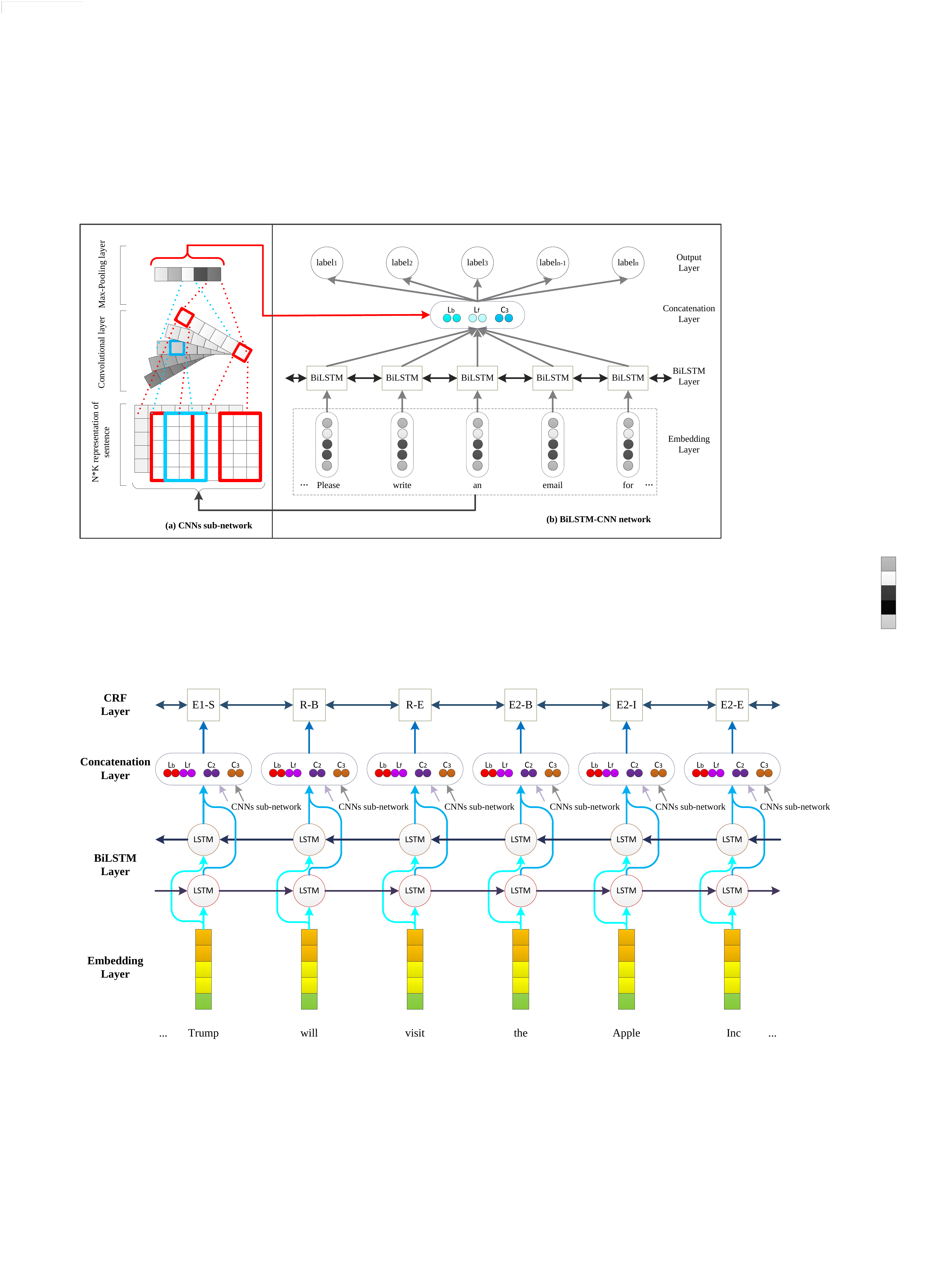}
	\caption{The structure of the hybrid neural network classification model.}
	\label{fig1}
\end{figure*}

\section{Related work}

User service intention identification is intended to judge what the consuming intents that are described by the users, such as booking a train ticket, order a hotel. Because of the short time in the intelligent service industry, the related technology is not mature. But it is gradually attracting the extensive attention of the academic and industrial community~\cite{Zhang2014Mining}.

The traditional intention classification works used more complicated artificial features, such as syntax information~\cite{Steedman2000Information,L2009A}, prosodic information~\cite{Weber2002Using}, lexical information~\cite{haffner2003optimizing,chelba2003speech}. The classification models adopted traditional statistical learning methods, including Bayesian~\cite{zhang2017first}, Decision tree~\cite{Xiao2011Research}, Support vector machine~\cite{zhang2017first},~\cite{qian2017identification} and soon. For example, Dai et al.~\cite{dai2006detecting} designed binary classifiers based on keywords and tags features in web pages to identify online business intentions. Yang et al.~\cite{qian2017identification} trained the classifier to identify the user's consumption intention by using the unigram features in Twitter, the lexical semantic features and verb features in WordNet. However, these methods had some limitations. First, manual processing of high-quality features was time-consuming and laborious. Additionally, the deep semantic information of the text was hard to learn by the way of building features so that the effect of classification is undesirable.

With the prevalence of deep learning recently, text classification based on the neural network has achieved good results. Reference~\cite{zhang2017first} used convolutional neural networks to automatically learn features for user intention recognition. And it also used LSTM networks to learn the long-distance dependencies of word sequences in text for intention recognition and achieves good results. Ding et al.~\cite{ding2015mining} adopted the domain adaptive neural network model to make binary categories to determine whether it contained user consumption intention.

\begin{table*}[h]
	\centering
	\caption{An example of user intent with category information.}
	\label{table3}
	\setlength{\tabcolsep}{3pt}
	\begin{center}
		
		\begin{tabular}{p{250pt}p{200pt}}	

			\specialrule{0.15em}{5pt}{5pt}
			Input message
			&Intent category
			\\ 	\specialrule{0.06em}{5pt}{5pt}
			Hello, nice to meet you!
			& Chit-chat
			\\ \specialrule{0.08em}{5pt}{5pt}
			I want to book an air ticket to Beijing.
			&	Task-oriented dialogue (Booking air tickets)
			\\ \specialrule{0.08em}{5pt}{5pt}
			I want to book a neat and low-priced inn near Wudaokou.
			&Task-oriented dialogue (Booking hotels)	\\ 
			\specialrule{0.15em}{5pt}{5pt}
			
		\end{tabular}
		\label{tab3}
	\end{center}
\end{table*}

\begin{table*}
	\centering
	\caption{The statistics of the released data for corpus.}
	\label{table4}
	\setlength{\tabcolsep}{3pt}
	\begin{center}
		
		\begin{tabular}{p{40pt}*{11}{p{36pt}}}
			\specialrule{0.15em}{5pt}{5pt}
			
			&app&bus&	calc&chat&cinemas&contacts&cookbook&datetime&email&epg&flight\\
			\specialrule{0.08em}{2pt}{2pt}
			\emph{Train} &36&24 &24&456&24&30&269 &18&24&107&62\\\specialrule{0.em}{2pt}{2pt}
			\emph{Dev} &18&8&8&114&10&10&88 &6&8&36&21\\\specialrule{0.em}{2pt}{2pt}
			\emph{Test} &18&8&8&50&8&10&90&6&8&36&21\\\specialrule{0.em}{2pt}{2pt}
			\emph{Sum}
			& 72&40&40&662&42&50&447&30&40&179&104\\ \specialrule{0.em}{2pt}{2pt}
			\specialrule{0.08em}{2pt}{5pt}
			
			&health&lottery&	map&match&message&music&news&novel&poetry&radio&riddle\\
			\specialrule{0.08em}{2pt}{2pt}
			\emph{Train} &55 &24&68 &24&63&66&58 &24&402&24&34\\\specialrule{0.em}{2pt}{2pt}
			\emph{Dev} &19 &8&23&8&21&22&19 &8&34&8&11\\\specialrule{0.em}{2pt}{2pt}
			\emph{Test}&18&8&24&8&21&22&19 &8&34&8&11\\\specialrule{0.em}{2pt}{2pt}
			\emph{Sum}&92&40&114&40&105&110&96&40&470&40&56\\	\specialrule{0.08em}{2pt}{5pt}
			
			&schedule
&stock& telephone
&train& 	translation&
tvchannel& video&
weather&website
&Total&\\
			\specialrule{0.08em}{2pt}{2pt}
			\emph{Train} &29&71&63&70&61&71&182&66&54&2583&\\\specialrule{0.em}{2pt}{2pt}
			\emph{Dev} &9&24&21&23&21&23&60&22&18&729&\\\specialrule{0.em}{2pt}{2pt}
			\emph{Test} &10&24&21&23&20&24&61&22&18&688&\\\specialrule{0.em}{2pt}{2pt}
			\emph{Sum}
&48&119&105&116&102&118&303&110&90&4000&\\\specialrule{0.15em}{5pt}{5pt}
			
		\end{tabular}
		\label{tab4}
	\end{center}
\end{table*}

\begin{table*}
	\centering
	\caption{The performances of our models working on each subclass the test set.}
	\label{table2}
	\setlength{\tabcolsep}{3pt}
	\begin{center}
		
		\begin{tabular}{p{55pt}*{8}{p{50pt}}}
			\specialrule{0.15em}{5pt}{5pt}
			
			\emph{Classification}&	app&	bus&	calc&	chat&	cinemas&	contacts&	cookbook&	datetime\\ 	\specialrule{0.08em}{5pt}{5pt}
			\emph{Precision}&	0.667	&1.0&	1.0	&0.922&	0.875&	1.0	&0.989&	0.667\\ 
			\specialrule{0.08em}{5pt}{5pt}
			\emph{Classification}&email&	epg&	radio&flight&	health&riddle&schedule&	stock\\ 
			\specialrule{0.08em}{5pt}{5pt}
			\emph{Precision}&1.0	&0.917&0.875&	1.0	&1.0&1.0&	0.900&	0.917\\ 
			\specialrule{0.08em}{5pt}{5pt}
			\emph{Classification}&lottery&	map	&match	&message	&music&	news&novel&	poetry\\ 
			\specialrule{0.08em}{5pt}{5pt}
			\emph{Precision}&	1.0	&1.0&	1.0	&1.0&	1.0	&1.0&	1.0	&1.0\\ 
			\specialrule{0.08em}{5pt}{5pt}
			\emph{Classification}&	telephone&	train&	translation&	tvchannel&	weather&website&video&avg\\
			\specialrule{0.08em}{5pt}{5pt}
			\emph{Precision}&	1.0&	1.0	&1.0&	0.875&	1.0&	0.833&	0.803&	0.943\\ 
			\specialrule{0.15em}{5pt}{5pt}
		\end{tabular}
		\label{tab2}
	\end{center}
\end{table*}

\section{Model}

This paper presents a hybrid neural network classification model. It uses BiLSTM and CNN to learn semantic information, and then uses the Softmax layer to output prediction tags. The model structure is shown in Figure~\ref{fig1}.

LSTM is a variant of the recurrent neural network (RNN), which is good at capturing long-range dependencies. 
The LSTM architecture consists of a set of recurrently connected subnets, known as memory cells, which is used to compute the current hidden vector $h_{t}$  based on the previous hidden vector  $h_{t-1}$, the previous cell vector $c_{t-1}$  and the current input embedding  $s_{t}$. Each time-step is a LSTM memory cell, which is implemented as the follows:
\begin{equation}
\left\{\begin{matrix}
i_{t}=\sigma (W_{si}s_{t}+W_{hi}h_{t-1}+W_{ci}c_{t-1}+b_{i}) \\ 
f_{t}=\sigma (W_{sf}s_{t}+W_{hf}h_{t-1}+W_{cf}c_{t-1}+b_{f}) \\ 
c_{t}=f_{t}c_{t-1}+i_{t}tanh(W_{sc}s_{t}+W_{hc}h_{t-1}+b_{c}) \\ 
o_{t}=\sigma (W_{so}s_{t}+W_{ho}h_{t-1}+W_{co}c_{t}+b_{o}) \\ 
h_{t}=o_{t}tanh(c_{t})
\end{matrix}\right.
\end{equation}
where, $i$, $f$,  $o$ and $c$ are the input gate, forget gate, output gate and cell vectors, respectively, $b$ is the bias,  $\sigma$ is the logistic sigmoid function, and $W$  are the trainable parameter matrixes.

For a given sentence \(\mathbf{X}=(\mathbf{x}_{1}, \mathbf{x}_{2}, . . ., \mathbf{x}_{n})\), the LSTM returns another representation \(\mathbf{H}=\left ( \mathbf{h}_{1}, \mathbf{h}_{2}, . . . , \mathbf{h}_{n} \right )\) about the sequence in the input. 
BiLSTM can present each sequence forwards and backwards to two separate hidden states to capture past (left) and future (right) information, respectively. BiLSTM is good at grasping the temporal semantics of the input sequence. It is useful for all kinds of text classification tasks.

Convolutional neural network (CNN) is a natural means of capturing salient local features from the whole sequence. Convolution is an operation between a vector of weights and a vector of the input sequence. The weights matrix is regarded as the filter for the convolution. In the example shown in Figure~\ref{fig1}(a), the input sequence consists of \emph{n} words and each word is mapped to \(k\)-dimension embedding representation. A convolutional filter is a list of linear layers with shared parameters. After feeding the output of a convolutional filter to a Max-Pooling layer, an output vector with fixed length can be obtained.

Let \(\mathbf{X}\)  denotes the sequence of information embeddings for every word in a sentence. The \(\mathbf{X}\)  is given as input to a BiLSTM, which returns a representation \(\left [ \mathbf{L}_{f},\mathbf{L}_{b} \right ]\) of the forward and backward context for each word. Then, The \(\mathbf{X}\) is given to the convolutional sub-network. The local information will be vectorization represent as \(\mathbf{C}_{3}\), by using convolutional filters with widths of 3. Next, we concatenate the bidirectional temporal features:  \(\mathbf{L}_{f}\) and \(\mathbf{L}_{b}\), and local spatial features:  \(\mathbf{C}_{3}\), as a single vector \(\mathbf{M} = \left [\mathbf{L}_{f}, \mathbf{L}_{b}, \mathbf{C}_{3}\right]\).  In the end, we use a dense layer with the Softmax activation function to yield the final predictions for every sentence. 

\section{Experiments}

\subsection{Experimental Setting}

We use the data provided by the SMP2017 Conference\footnote{http://ir.hit.edu.cn/SMP2017-ECDT} as the experimental data. There are two top categories, namely, chit-chat and task-oriented dialogue. The task-oriented dialogue also includes 30 sub categories, such as ``cookbook'', ``music'', ``bus'', etc. In this evaluation, we only consider to classify the user intent in single utterance. An example of user intent with category information as shown in Table~\ref{table3} and the statistics of the corpus is shown in Table~\ref{table4}..

We implement the neural network using the Keras library\footnote{https://keras.io/}. The training set, validation set, and the test set contain 2583, 729 and 688 records, respectively. The batch size is fixed to 10. We use early stopping based on performance on the validation set. The number of LSTM units is 50 and the number of feature maps for each convolutional filter is 50. Parameter optimization is performed with Adam optimizer. Here, an initial learning rate is 0.001, and the learning rate is reduced by a factor of 0.1 if no improvement of the loss function is seen for some epochs. Besides, to mitigate over-fitting, we apply the dropout method to regularize our model.

\subsection{Experimental Results}

 We tested the performances of our model on each subclass, as shown in Table~\ref{table2}. Our model obtains 1.0 accuracy on 19 subclasses, and the recognition accuracy is above 0.9 in 24 subclasses. There is poor recognition in some categories, such as ``app” and ``datetime”. This may be due to the uneven distribution of data between categories in the dataset, and some categories are easily confused and difficult to distinguish. 

\begin{table}[h]
	\caption{The results of various models working on the test set.}
	\label{table1}
	\begin{center}
		
		\begin{tabular}{p{48pt}*{3}{p{48pt}}}
			\specialrule{0.15em}{5pt}{5pt}
			\emph{Models}&	Naive Bayes	&Decision Tree&	SVM\\ \specialrule{0.08em}{5pt}{5pt}
			\emph{F1 score}&	0.835&	0.809&	0.895\\ \specialrule{0.08em}{5pt}{5pt}
			CNN	&LSTM&	Model in~\cite{zhang2017first}&	Our Model \\ \specialrule{0.08em}{5pt}{5pt}
			0.909&	0.901&	0.939&\textbf{0.942}\\
			\specialrule{0.15em}{5pt}{5pt}
		\end{tabular}
		\label{tab1}
	\end{center}
\end{table}

In addition, we report the results of various models working on the test set. We can get some conclusions as follows from the Table~\ref{table1}. The method based on SVM has achieved better results than other traditional classification models, including Naive Bayes or Decision Tree based. However, the model based on the neural network has a better effect than the traditional machine learning classification models. The model based on CNN is better than that based on LSTM. It may be caused by that the sentences in this dataset are short, and the sequence order clutter caused by the colloquial problem is serious, these reasons are bad for the recurrent neural network to capture the temporal semantics. In addition, our model combines the advantages of LSTM and CNN to achieve better results than all the other models, including the best model was shown in the reference~\cite{zhang2017first}.

\section{Conclusion}

In view of the poor performances of the intelligent service system to identify the user's service intentions, we propose a hybrid neural network model based on BiLSTM and CNN, which can automatically learn the semantic features of the text, without artificial intervention. It achieves better results than the existing models on the open test set. And we believed that our model will achieve better results as the number of tagged data increases. In future work, we will study more efficient models to further improve the efficiency of recognition. For example, the attention mechanism is added to the neural network model to highlight the influence of key text words on recognition.


\section*{Acknowledgment}

The authors would like to thank the reviewers for their comments on this article.
This work was supported in part by the National Natural Science Foundation of China under Grant 71571136, in part by the National Basic Research Program of China under Grant 2014CB340404, and in part by the Project of Science and Technology Commission of Shanghai Municipality under Grant 16JC1403000 and Grant 14511108002.



%

%
%

\bibliographystyle{IEEEtran}

\bibliography{umi}

\begin{thebibliography}{10}
\providecommand{\url}[1]{#1}
\csname url@samestyle\endcsname
\providecommand{\newblock}{\relax}
\providecommand{\bibinfo}[2]{#2}
\providecommand{\BIBentrySTDinterwordspacing}{\spaceskip=0pt\relax}
\providecommand{\BIBentryALTinterwordstretchfactor}{4}
\providecommand{\BIBentryALTinterwordspacing}{\spaceskip=\fontdimen2\font plus
\BIBentryALTinterwordstretchfactor\fontdimen3\font minus
  \fontdimen4\font\relax}
\providecommand{\BIBforeignlanguage}[2]{{%
\expandafter\ifx\csname l@#1\endcsname\relax
\typeout{** WARNING: IEEEtran.bst: No hyphenation pattern has been}%
\typeout{** loaded for the language `#1'. Using the pattern for}%
\typeout{** the default language instead.}%
\else
\language=\csname l@#1\endcsname
\fi
#2}}
\providecommand{\BIBdecl}{\relax}
\BIBdecl

\bibitem{Liu2018The}
B.~Liu and Y.~Wang, ``The application of deep learning technology to
  intelligent service,'' \emph{Research on Library Science}, 2018.

\bibitem{Poniszewska2018Endowing}
A.~Poniszewska-Maranda, D.~Kaczmarek, N.~Kryvinska, and F.~Xhafa,
  \emph{Endowing IoT Devices with Intelligent Services}, 2018.

\bibitem{Fubo2015Consumption}
B.~Fu and L.~Ting, ``Consumption intent recognition for social media: task,
  challenge and opportunity,'' \emph{Intelligent Computer and Applications},
  vol.~5, no.~4, pp. 1--4, 2015.

\bibitem{Zhang2014Mining}
F.~Zhang, N.~J. Yuan, D.~Lian, and X.~Xie, ``Mining novelty-seeking trait
  across heterogeneous domains,'' in \emph{International Conference on World
  Wide Web}, 2014, pp. 373--384.

\bibitem{haffner2003optimizing}
P.~Haffner, G.~Tur, and J.~H. Wright, ``Optimizing svms for complex call
  classification,'' in \emph{Acoustics, Speech, and Signal Processing, 2003.
  Proceedings.(ICASSP'03). 2003 IEEE International Conference on},
  vol.~1.\hskip 1em plus 0.5em minus 0.4em\relax IEEE, 2003, pp.
  I--632--I--635.

\bibitem{Steedman2000Information}
M.~Steedman, ``Information structure and the syntax-phonology interface,''
  \emph{Linguistic Inquiry}, vol.~31, no.~4, pp. 649--689, 2000.

\bibitem{L2009A}
L.~López, \emph{A derivational syntax for information structure}.\hskip 1em
  plus 0.5em minus 0.4em\relax Oxford University Press, 2009.

\bibitem{Weber2002Using}
F.~Weber, L.~Manganaro, B.~Peskin, and E.~Shriberg, ``Using prosodic and
  lexical information for speaker identification,'' in \emph{IEEE International
  Conference on Acoustics, Speech, and Signal Processing}, 2002, pp.
  I--141--I--144.

\bibitem{chelba2003speech}
C.~Chelba, M.~Mahajan, and A.~Acero, ``Speech utterance classification,'' in
  \emph{Acoustics, Speech, and Signal Processing, 2003.
  Proceedings.(ICASSP'03). 2003 IEEE International Conference on},
  vol.~1.\hskip 1em plus 0.5em minus 0.4em\relax IEEE, 2003, pp.
  I--280--I--283.

\bibitem{zhang2017first}
W.-N. Zhang, Z.~Chen, W.~Che, G.~Hu, and T.~Liu, ``The first evaluation of
  chinese human-computer dialogue technology,'' \emph{arXiv preprint
  arXiv:1709.10217}, 2017.

\bibitem{Xiao2011Research}
T.~Xiao, ``Research on user query intention recognition technology based on
  decision algorithm,'' Ph.D. dissertation, South China Normal University,
  2011.

\bibitem{qian2017identification}
Y.~QIAN, X.~DING, T.~LIU, and Y.~CHEN, ``Identification method of user's travel
  consumption intention in chatting robot,'' \emph{SCIENTIA SINICA
  Informationis}, vol.~47, no.~8, p. 997, 2017.

\bibitem{dai2006detecting}
H.~K. Dai, L.~Zhao, Z.~Nie, J.-R. Wen, L.~Wang, and Y.~Li, ``Detecting online
  commercial intention (oci),'' in \emph{Proceedings of the 15th international
  conference on World Wide Web}.\hskip 1em plus 0.5em minus 0.4em\relax ACM,
  2006, pp. 829--837.

\bibitem{ding2015mining}
X.~Ding, T.~Liu, J.~Duan, and J.-Y. Nie, ``Mining user consumption intention
  from social media using domain adaptive convolutional neural network.'' in
  \emph{AAAI}, vol.~15, 2015, pp. 2389--2395.

\end{thebibliography}

\end{document}